\documentclass[10pt,twocolumn,letterpaper]{article}

\usepackage{cvpr}
\usepackage{times}
\usepackage{epsfig}
\usepackage{graphicx}
\usepackage{amsmath}
\usepackage{amssymb}

\usepackage{subfig,multirow,multicol,array}
\usepackage{amsfonts,amssymb}
\usepackage{algorithmic}
\usepackage[ruled]{algorithm2e}	

\usepackage[breaklinks=true,bookmarks=false]{hyperref}

\cvprfinalcopy 

\def\cvprPaperID{3970} 

\DeclareMathOperator*{\argmin}{arg\,min}

\usepackage{colortbl}
\definecolor{mygray}{gray}{.9}

\ifcvprfinal\fi
\begin{document}

\title{Robust Video Content Alignment and Compensation\\for Rain Removal in a CNN Framework}

\author{Jie Chen$^1$, Cheen-Hau Tan$^1$, Junhui Hou$^2$, Lap-Pui Chau$^1$, and He Li$^3$\\
\small $^1$School of Electrical and Electronic Engineering, Nanyang Technological University, Singapore\\
\small $^2$Department of Computer Science, City University of Hong Kong\\
\small $^3$Singapore Technologies Dynamics Pte Ltd\\
{\tt\small \{chen.jie, cheenhau, elpchau\}@ntu.edu.sg, jh.hou@cityu.edu.hk, lihe@stengg.com}
}

\maketitle

\begin{abstract}
	
Rain removal is important for improving the robustness of outdoor vision based systems. Current rain removal methods show limitations either for complex dynamic scenes shot from fast moving cameras, or under torrential rain fall with opaque occlusions. 
We propose a novel derain algorithm, which applies superpixel (SP) segmentation to decompose the scene into depth consistent units. 
Alignment of scene contents are done at the SP level, which proves to be robust towards rain occlusion and fast camera motion.
Two alignment output tensors, i.e., optimal temporal match tensor and sorted spatial-temporal match tensor, provide informative clues for rain streak location and occluded background contents to generate an intermediate derain output.
These tensors will be subsequently prepared as input features for a convolutional neural network to restore high frequency details to the intermediate output for compensation of mis-alignment blur. 
Extensive evaluations show that up to 5\textit{dB} reconstruction PSNR advantage is achieved over state-of-the-art methods. Visual inspection shows that much cleaner rain removal is achieved especially for highly dynamic scenes with heavy and opaque rainfall from a fast moving camera. 
\end{abstract}\vspace{-0.2cm}

\section{Introduction}\label{sec_introduction}

Modern intelligent systems rely more and more on visual information as input.
However, in an outdoors setting, visual input quality and in turn, system performance, could be seriously degraded by atmospheric turbulences \cite{Garg2007,narasimhan2002vision}.
One such turbulence, rain streaks, degrade image contrast and visibility, obscure scene features, and could be misconstrued as scene motion by computer vision algorithms.
Rain removal is therefore vital to ensure the robustness of outdoor vision-based systems.

\begin{figure}
	\centering
	\includegraphics[width=0.95\linewidth]{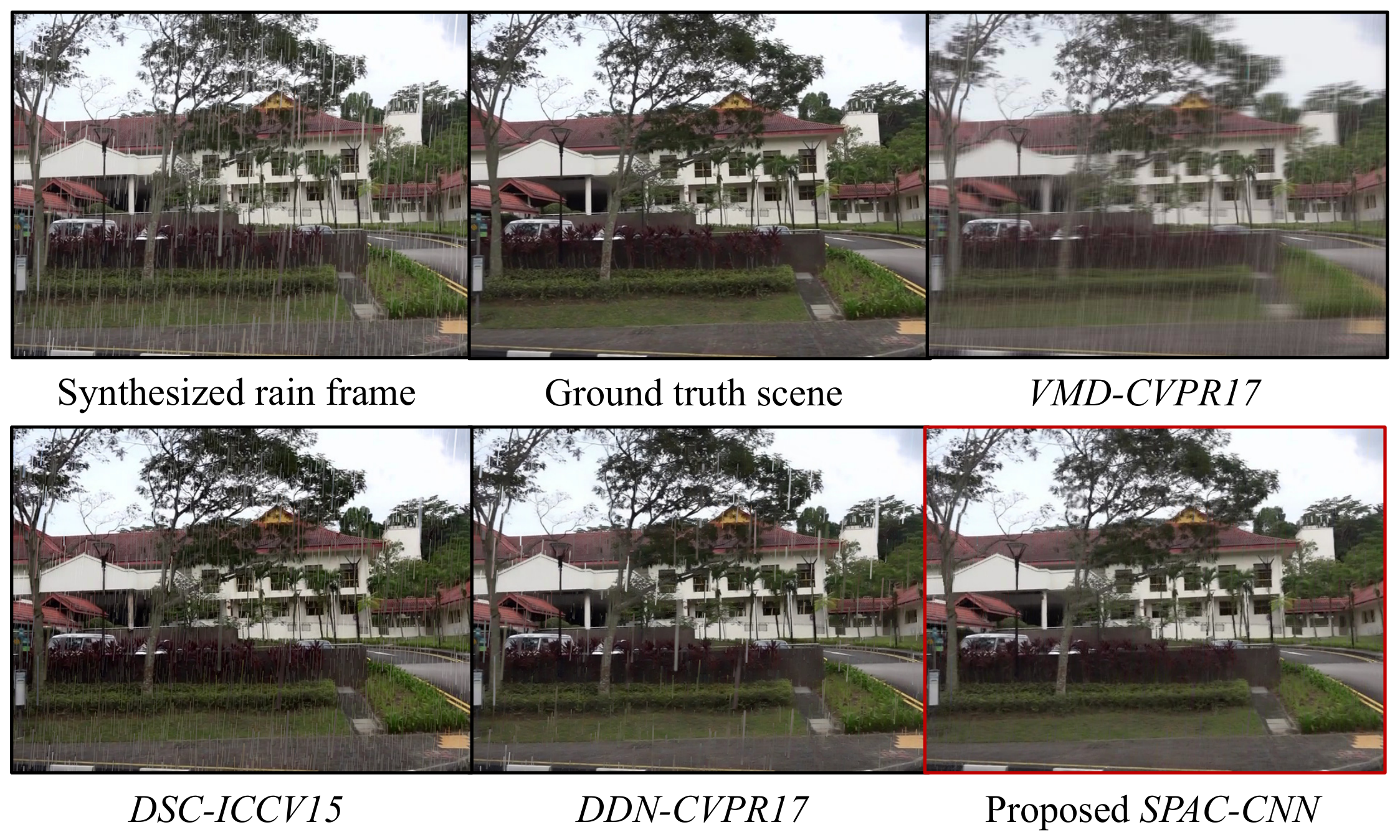}
	\caption{Comparison of derain outputs by different algorithms for a challenging video sequence with fast camera motion and heavy rain fall. Image-based derain methods, i.e., discriminative sparse coding (\textit{DSC}) \cite{luo2015removing} and deep detail network (\textit{DDN}) \cite{fu2017removing} fail to remove large and/or opaque rain streaks. A video-based method via matrix decomposition (\textit{VMD}) \cite{ren2017video} creates serious blur due to fast camera motion. Our proposed \textit{SPAC-CNN} can cleanly remove the rain streaks and preserve scene contents truthfully.}
	\vspace{-0.4cm}
	\label{fig_teaserComp}
\end{figure}

\begin{figure*}
	\begin{center}
		\includegraphics[width=0.9\linewidth]{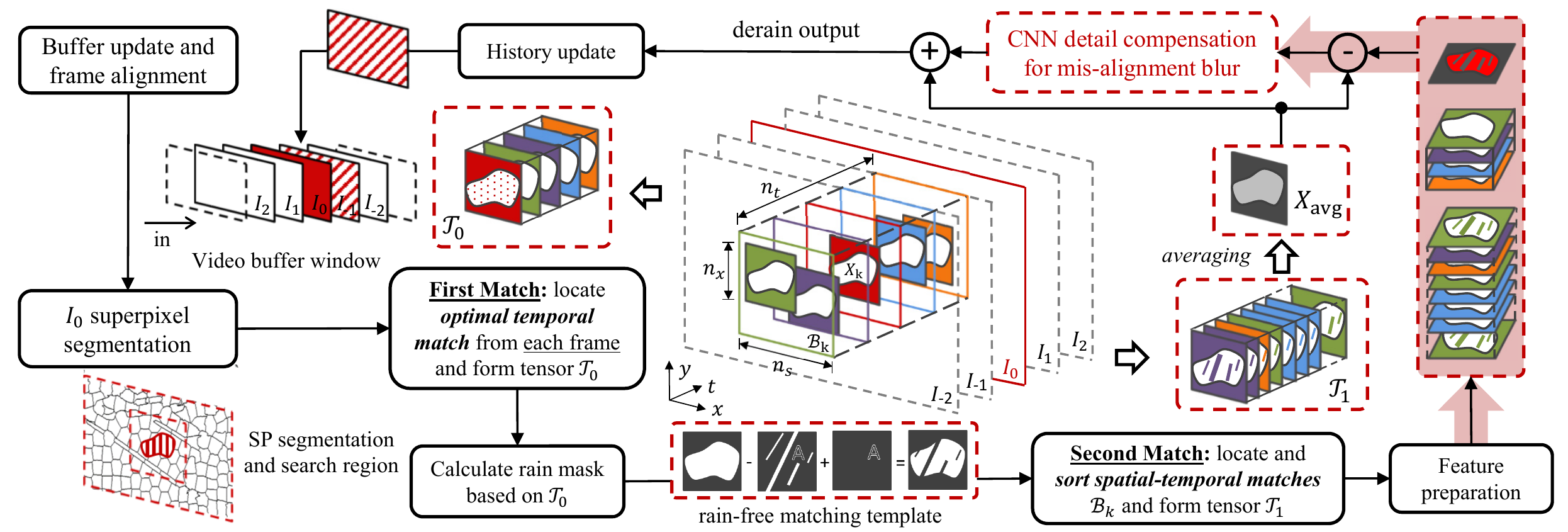}
	\end{center}
	\vspace{-0.5cm}
	\caption{System diagram for the proposed \textit{SPAC-CNN} rain removal algorithm.}
	\vspace{-0.5cm}
	\label{fig_system}
\end{figure*}

There are two categories of methods for rain removal -- image-based methods, which rely solely on the information of the processed frame, and video-based methods, which also utilize temporal clues from neighboring frames.
Due to the lack of temporal information, image-based methods face difficulties in recovering from torrential rain with large and opaque occlusions.

To properly utilize temporal information, video-based methods require scene content to be aligned throughout consecutive frames.
However, this requirement is challenging due to two factors -- motion of the camera and dynamic scene content, i.e., presence of moving object.
Previous works tackle these two issues separately.
Camera motion-induced scene content shifts can be reversed using global frame alignment \cite{Tripathi2012, tan2014dynamic}.
However, the granularity of global alignment is too large when scene depth range is large; 
parts of scene content will be poorly aligned.
Scene content shifts due to scene object motions could cause the scene to be misclassified as rain. 
One solution is to identify and exclude these pixels.
This approach, however, is unable to remove rain that overlaps moving objects. 

In this paper, we propose a novel and elegant framework that simultaneously solves both issues for the video-based approach -- rain removal based on robust SuperPixel (SP) Alignment between video frames followed by detail Compensation in a CNN framework (\textit{SPAC-CNN}). 
First, the target video frame is segmented into SPs and each SP is aligned with its temporal neighbors. 
This step simultaneously aligns both the scene background and moving objects without prior assumptions about moving objects. 
Scene content is also much better aligned at a SP level granularity.
An intermediate derain output can be obtained by averaging the aligned SPs, which unavoidably introduces blurring.
We restore the rain free details to the intermediate output by extracting the information from the aligned SPs using a convolutional neural network (CNN).

Extensive experiments show that our proposed algorithm achieves up to 5\textit{dB} reconstruction advantage over state-of-the-art rain removal methods. Visual inspection shows that rain is much better removed, especially for heavy and opaque rainfall regions over highly dynamic scene content. 
Fig.~\ref{fig_teaserComp} illustrates the advantage of our proposed algorithm over existing methods in a challenging video sequence.
The contribution of this work can be generalized as follows: \vspace{-0.2cm}

\begin{enumerate}
	\item We propose a novel spatial-temporal content alignment algorithm at SP level, which can handle fast camera motion and dynamic scene contents in one framework. This mechanism greatly outperforms existing scene motion analysis methods that models background and foreground motion separately. \vspace{-0.2cm}
	\item The strong local properties of SPs can robustly counter heavy rain interferences, and facilitate much more accurate alignment. Owing to such robust alignment, accurate temporal correspondence could be established for rain occlusions such that heavily occluded backgrounds could be truthfully restored. This greatly outperforms image-based derain methods in which recovery of large and opaque rain occlusions remain the biggest challenge. \vspace{-0.2cm}
	\item We propose a set of very efficient spatial-temporal features for the compensation of high frequency details lost during the deraining process. An efficient CNN network is designed, and a synthetic rain video dataset is created for training the CNN. \vspace{-0.2cm}
\end{enumerate}


\section{Related Work}


Rain removal based on a single image is intrinsically a challenging one, since it only relies on visual features and priors to distinguish rain from the background.
Local photometric, geometric, and statistical properties of rain have been studied in \cite{Garg2007, Garg2005When, Zheng2013single, kim2013single-image}.
Li et al. \cite{li2016rain} models background and rain streaks as layers to be separated.
Under the sparse coding framework, rain and backgrounds can be efficiently separated either with classified dictionary atoms \cite{kang2012automatic,chen2014visual}, or via discriminative sparse coding \cite{luo2015removing}.
Convolutional Neural Networks have been very effective in both high-level vision tasks \cite{krizhevsky2012imagenet} and low-level vision applications for capturing signal characteristics \cite{Kim2016accurate,zhang2017beyong}.
Hence, different network structures and features were explored for rain removal, such as the deep detail network \cite{fu2017removing}, and the joint rain detection and removal model \cite{yang2017deep}.
Due to the lack of temporal information, heavy and opaque rain is difficult to be distinguished from scene structures. Full recovery of a seriously occluded scene is almost impossible.


The temporal information from a video sequence provides huge advantage for rain removal \cite{Garg2004,Barnum2009Frequency,Santhaseelan2012Phase,santhaseelan2015utilizing,you2016adherent}. True rain pixels are separated from moving object pixels based on statistics of intensity values \cite{Tripathi2012} or chromatic values \cite{zhang2006}, on geometric properties of connected candidate pixels \cite{bossu2011rain}, or on segmented motion regions \cite{chen2014rain}.
Kim's work \cite{kim2015video} compensates for scene content motion by using optical flow for content alignment.
Ren et al. \cite{ren2017video} decomposes a video into background, rain, and moving objects using matrix decomposition. Moving objects are derained by temporally aligning them using patch matching, while the moving camera effect is modeled using a frame transform variable.
Temporal derain methods can handle occlusions much better than image-based methods; however, these methods perform poorly for complex dynamic scenes shot from fast moving cameras.


\section{Proposed Model} \label{sec_proposedModel}

Throughout the paper, scalars are denoted by italic lower-case letters, 2D matrices by upper-case letters, 3D tensors, functions, and operators by script letters.

Given a target derain video frame $I_0$, we look at its immediate past and future neighbor frames to create a sliding buffer window of length $n_t$: 
$\{I_i|i=[-\frac{n_t-1}{2},\frac{n_t-1}{2}]\}$. Here, negative and positive $i$ indicate past and future frames, respectively. We only derain the Y luminance channel. The derain output is used to update the history buffer (Fig. \ref{fig_system}). Such history update mechanism ensures cleaner derain for heavy rainfall scenarios.

The system diagram for the proposed \textit{SPAC-CNN} rain removal algorithm is shown in Fig. \ref{fig_system}. The algorithm can be divided into two parts: first, video content alignment is carried out at SP level, which consists two SP template matching operations that produce two output tensors: the \textbf{\textit{optimal temporal match tensor}} $\mathcal{T}_0$, and the \textbf{\textit{sorted spatial-temporal match tensor}} $\mathcal{T}_1$. An intermediate derain output $X_\text{avg}$ is calculated by averaging the \textit{slices}\footnote{A slice is a two-dimensional section of a higher dimensional tensor, defined by fixing all but two indices \cite{kolda2009tensor}.} of the tensor $\mathcal{T}_1$. 
Second, these two tensors will be prepared as input features to a CNN to compensate the high frequency details lost in $X_\text{avg}$ caused by mis-alignment blur. The detail of each component will be explained in this section.

\subsection{Robust Content Alignment via Superpixel Spatial-Temporal Matching}

One of the most important procedure for video-based derain algorithms is the estimation of content correspondence between video frames. With accurate content alignment, rain occlusions could be easily detected and removed with information from the temporal axis.

\subsubsection{Content Alignment: Global vs. Superpixel} \label{sec_alignment}

The popular solution to compensate camera motion between two frames is via a homography transform matrix estimated based on global consensus of a group of matched feature points \cite{bay2008speeded, torr2000mlesac}. Due to the reasons analyzed in Sec. \ref{sec_introduction}, perfect content alignment can never be achieved for all pixels with a global transform at whole frame level, especially for dynamic scenes with large depth range.

The solution naturally turns to pixel-level alignment, which faces no fewer challenges: first, feature points are sparse, and feature-less regions are difficult to align; more importantly, rain streak occlusions will cause serious interferences to feature matching at single pixel level. Information from larger areas are required to overcome rain interferences. This lead us to our final solution: to decompose images into smaller depth consistent units.

\begin{figure}[t]
	\begin{center}
		\includegraphics[width=0.76\linewidth]{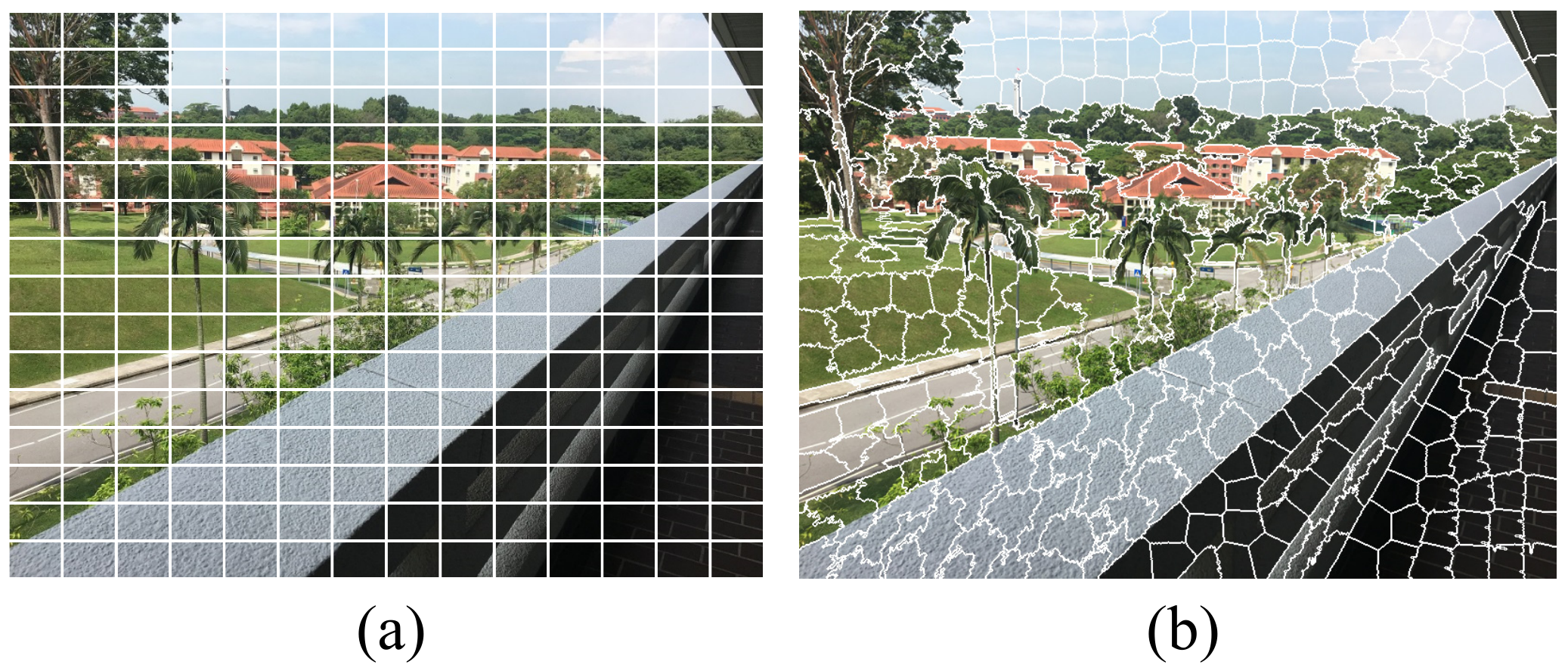}
	\end{center}\vspace{-0.5cm}
	\caption{Rectangular and SP segmentation units.}
	\label{fig_warpingOutput}
	\vspace{-0.5cm}
\end{figure}

The concept of SuperPixel (SP) is to group pixels into perceptually meaningful atomic regions \cite{achanta2012slic,bergh2012seeds,li2015superpixel}. Boundaries of SP usually coincide with those of the scene contents. Comparing Fig. \ref{fig_warpingOutput}(a) and (b), the SPs are very adaptive in shape, and are more likely to segment uniform depth regions compared with rectangular units.
We adopt SP as the basic unit for content alignment.

\subsubsection{Optimal Temporal Matching for Rain Detection}\label{sec_T0}
 
Let $\mathcal{P}_k$ denote the set of pixels that belong to the $k$-th SP on $I_0$. Let $X_k\in\mathbb{R}^{n_x\times n_x}$ be the bounding box that covers all pixels in $\mathcal{P}_k$ ($\mathcal{P}_k\subset X_k$). Let $\mathcal{B}_{k}\in\mathbb{R}^{n_\text{s}\times n_\text{s}\times n_t}$ denote a spatial-temporal buffer centered on $\mathcal{P}_k$. As illustrated in Fig. \ref{fig_system}, $\mathcal{B}_k$ spans the entire sliding buffer window, and its spatial range $n_s\times n_s$ is set to cover the possible motion range of $\mathcal{P}_k$ in its neighboring frames.

Pixels within the same SP are very likely to belong to the same object and possess identical motion between adjacent frames.
Therefore, we can approximate the SP appearance in their adjacent frames based on its appearance in the current frame via linear translations. 

\begin{figure}
	\centering
	\vspace{-0.5cm}
	\includegraphics[width=0.8\linewidth]{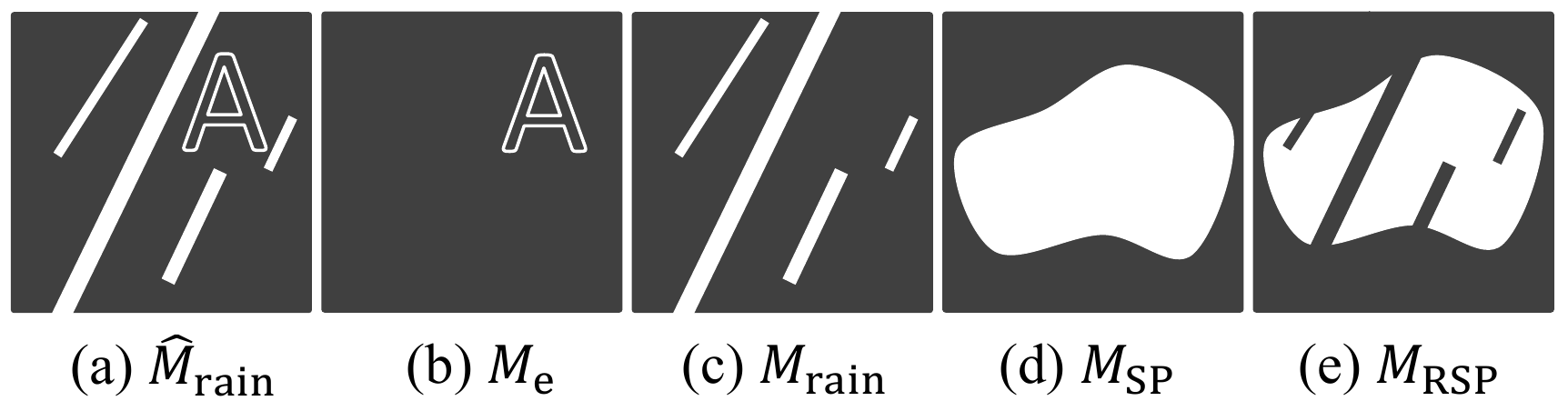}
	\caption{Illustration of various masks and matching templates used in the proposed algorithm.}
	\vspace{-0.5cm}
	\label{fig_maskMatch}
\end{figure}

Searching for the reference SP is done by template matching of the target SP at all candidate locations in $\mathcal{B}_k$. A match location is found at frame $I_{t'}$ according to: \vspace{-0.2cm}
\begin{align}
(\hat{u},\hat{v})=\argmin_{u,v}\sum_{(x,y)\in X_k}&|\mathcal{B}_k(x+u,y+v,t')\\\notag 
&-X_k(x,y)|^2\odot M_\text{SP}(x,y).
\end{align}
As shown in Fig.~\ref{fig_maskMatch}(d), $M_\text{SP}$ indicates SP pixels $\mathcal{P}_k$ in the bounding box $X_k$. $\odot$ denotes element-wise multiplication. Each match at a different frame becomes a \textit{temporal slice} for the \textbf{\textit{optimal temporal match tensor}} $\mathcal{T}_0\in \mathbb{R}^{n_x\times n_x\times n_t}$: \vspace{-0.2cm}
\begin{equation}
\mathcal{T}_0(\cdot,\cdot,t')= \mathcal{B}_k(x+\hat{u},y+\hat{v},t'),~(x,y)\in X_k.
\end{equation} \vspace{-0.6cm}

Based on the temporal clues provided by $\mathcal{T}_0$, a rain mask can be estimated. Since rain increases the intensity of its covered pixels \cite{Garg2004}, rain pixels in $X_k$ are expected to have higher intensity than their collocated temporal neighbors in $\mathcal{T}_0$. We first compute a binary tensor $\mathcal{M}_{\text{0}}\in\mathbb{R}^{n_x\times n_x\times n_t}$ to detect positive temporal fluctuations: \vspace{-0.2cm} 
\begin{align}
\mathcal{M}_{0}&=\begin{cases}
1& \mathcal{R}(X_k,n_t)- \mathcal{T}_0 \ge \epsilon_{\text{rain}}\\
0& \text{otherwise}
\end{cases},
\end{align}
where operator $\mathcal{R}(\Phi,\psi)$ is defined as replicating the 2D slices $\Phi\in\mathbb{R}^{n_1\times n_2}$ $\psi$ times and stacking along the thrid dimension into a tensor of $\mathbb{R}^{n_1\times n_2\times \psi}$. To robustly handle re-occurring rain streaks, we classify pixels as rain when at least 3 positive fluctuations are detected in $\mathcal{M}_0$. An initial rain mask $\hat{M}_{\text{rain}}\in\mathbb{R}^{n_x\times n_x}$ can be calculated as: \vspace{-0.2cm}
\begin{equation}
\hat{M}_{\text{rain}}(x,y)=[\sum_t \mathcal{M}_{\text{0}}(x,y,t)] \ge 3.
\end{equation}\vspace{-0.3cm}

Due to possible mis-alignment, edges of background could be misclassified as rain. Since rain steaks don't affect values in the chroma channels (Cb and Cr), a rain-free edge map $M_{\text{e}}$ could be calculated by thresholding the sum of gradients of these two channels with $\epsilon_\text{e}$. The final rain mask $M_{\text{rain}}\in\mathbb{R}^{n_\text{s}\times n_\text{s}}$ is calculated as:\vspace{-0.2cm}
\begin{equation}
M_{\text{rain}} = \hat{M}_{\text{rain}}\odot (1-M_e).
\end{equation}
A visual demonstration of $\hat{M}_{\text{rain}}$, $M_{\text{e}}$, and $M_{\text{rain}}$ is shown in Fig. \ref{fig_maskMatch}(a), (b), and (c), respectively. In our implementation, $\epsilon_\text{rain}$ is set to $0.012$ while $\epsilon_\text{edge}$ is set to $0.2$.\vspace{-0.2cm}

\subsubsection{Sorted Spatial-Temporal Template Matching for Rain Occlusion Suppression} \label{sec_T1}

The second round of template matching will be carried out based on the following cost function: \vspace{-0.2cm}
\begin{align}\label{eqn_T1cost}
E(u,v,t)= \sum_{(x,y)\in X_k}&|\mathcal{B}_k(x+u,y+v,t)\\ \notag  
- &X_k(x,y)|^2\odot M_{\text{RSP}}(x,y).
\end{align}
The \textbf{rain-free matching template} $M_
\text{RSP}$ is calculated as: \vspace{-0.2cm} 
\begin{equation}
M_{\text{RSP}}= M_{\text{SP}} \odot (1-M_{\text{rain}}).
\end{equation}
As shown in Fig. \ref{fig_maskMatch}(e), only the \textbf{\textit{rain-free background SP pixels}} will be used for matching.
Each candidate locations in $\mathcal{B}_k$ (except current frame $\mathcal{B}_k(\cdot,\cdot,0)$) are sorted in ascending order based on their cost $E$ defined in Eq. (\ref{eqn_T1cost}). The top $n_{st}$ candidates with smallest $E$ will be stacked as slices to form the \textbf{\textit{sorted spatial-temporal match tensor}} $\mathcal{T}_1\in\mathbb{R}^{n_\text{s}\times n_\text{s}\times n_{st}}$. 
 
The slices of $\mathcal{T}_1(\cdot,\cdot,t)$ are expected to be well-aligned to the current target SP $\mathcal{P}_k$, and is robust to interferences from the rain. 
Since rain pixels are temporally randomly and sparsely distributed within $\mathcal{T}_1$, when $n_{st}$ is sufficiently large, we can get a good estimation of the rain free image through \textit{tensor slice averaging}, which functions to suppress rain induced intensity fluctuations, and bring out the occluded background pixels: \vspace{-0.2cm}
\begin{equation}\label{eqn_Xavg}
X_\text{avg}=\frac{\sum_{t}\mathcal{T}_1(\cdot,\cdot,t)}{n_{st}}.
\end{equation}

\begin{figure}
	\centering
	\vspace{-0.5cm}
	\includegraphics[width=1\linewidth]{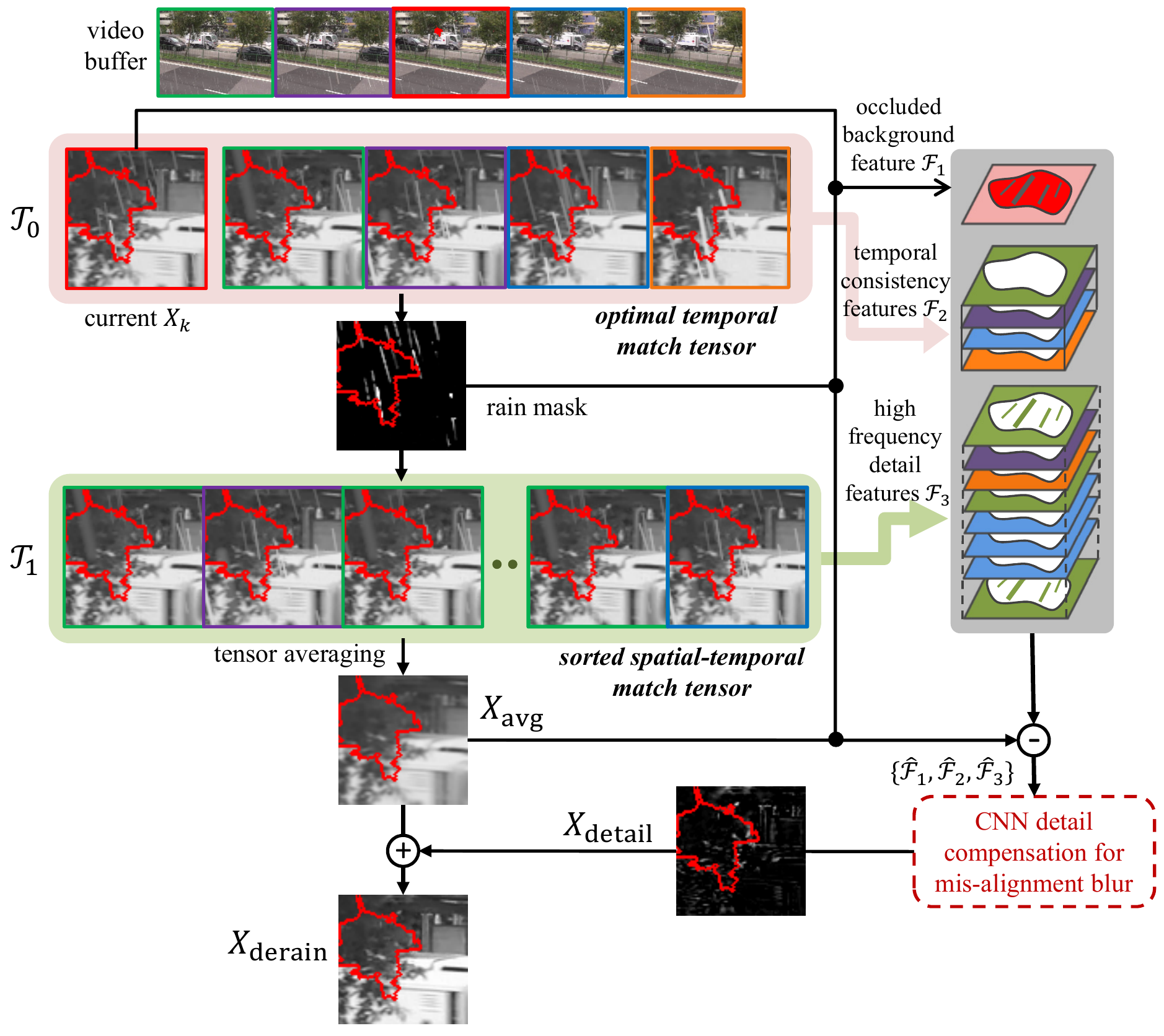}
	\caption{Illustration of feature preparation for the detail recovery CNN. }
	\vspace{-0.5cm}
	\label{fig_T0T1forming}
\end{figure}

Fig. \ref{fig_T0T1forming} gives a visual example of $X_\text{avg}$ and its calculation flow. We can see that all rain streaks have been suppressed in $X_\text{avg}$ after the averaging.

\subsection{Detail Compensation for Mis-Alignment Blur}

The averaging of $\mathcal{T}_1$ slices provides a good estimation of rain free image; however, it creates noticeable blur due to un-avoidable mis-alignment, especially when the camera motion is fast. To compensate the lost high frequency content details without reintroducing the rain streaks, we propose to use a CNN model for the task. 

\subsubsection{Occluded Background Feature} 

$X_\text{avg}$ from Eq. (\ref{eqn_Xavg}) can be used as one important clue to recover rain occluded pixels. Rain streak pixels indicated by the rain mask $M_\text{rain}$ are replaced with corresponding pixels from $X_\text{avg}$ to form the first feature $\mathcal{F}_1\in\mathbb{R}^{n_x\times n_x\times 1}$:\vspace{-0.2cm}
\begin{equation}\label{eqn_feature1}
\mathcal{F}_1= X_k\odot (1-M_\text{rain})+ X_\text{avg}\odot M_\text{rain}.
\end{equation}

Note that the feature $\mathcal{F}_1$ itself is already a reasonable derain output. However its quality is greatly limited by the correctness of the rain mask $M_\text{rain}$. For \textbf{false positive}\footnote{False positive rain pixels refer to background pixels falsely classified as rain; false negative rain pixels refer to rain pixels falsely classified as background.} rain pixels, $X_\text{avg}$ will introduce content detail loss; for \textbf{false negative} pixels, rain streaks will be added back from $X_k$. This calls for more informative features.

\subsubsection{Temporal Consistency Feature}
 
The temporal consistency feature is designed to handle \textbf{false negative} rain pixels in $M_\text{rain}$, which falsely add rain streaks back to $\mathcal{F}_1$. For a correctly classified and recovered pixel (a.k.a. \textbf{true positive}) in Eq. (\ref{eqn_feature1}), intensity consistency should hold such that for the collocated pixels in the neighboring frames, there are only positive intensity fluctuations caused by rain in \textit{those} frames. Any obvious negative intensity drop along the temporal axis is a strong indication that such pixel is a \textbf{false negative} pixel.

The temporal slices in $\mathcal{T}_0$ establishes optimal temporal correspondence at each frame, which embeds enough information for the CNN to deduce the above analyzed false negative logic, therefore they shall serve as the second feature $\mathcal{F}_2\in\mathbb{R}^{n_x\times n_x\times (n_t-1)}$:
\begin{equation}
\mathcal{F}_2= \{\mathcal{T}_0(\cdot,\cdot,t)|~t=[-\frac{n_t-1}{2},\frac{n_t-1}{2}],~t\neq 0\}.
\end{equation}\vspace{-0.5cm}

\begin{figure}[t]
	\begin{center}
		\includegraphics[width=1\linewidth]{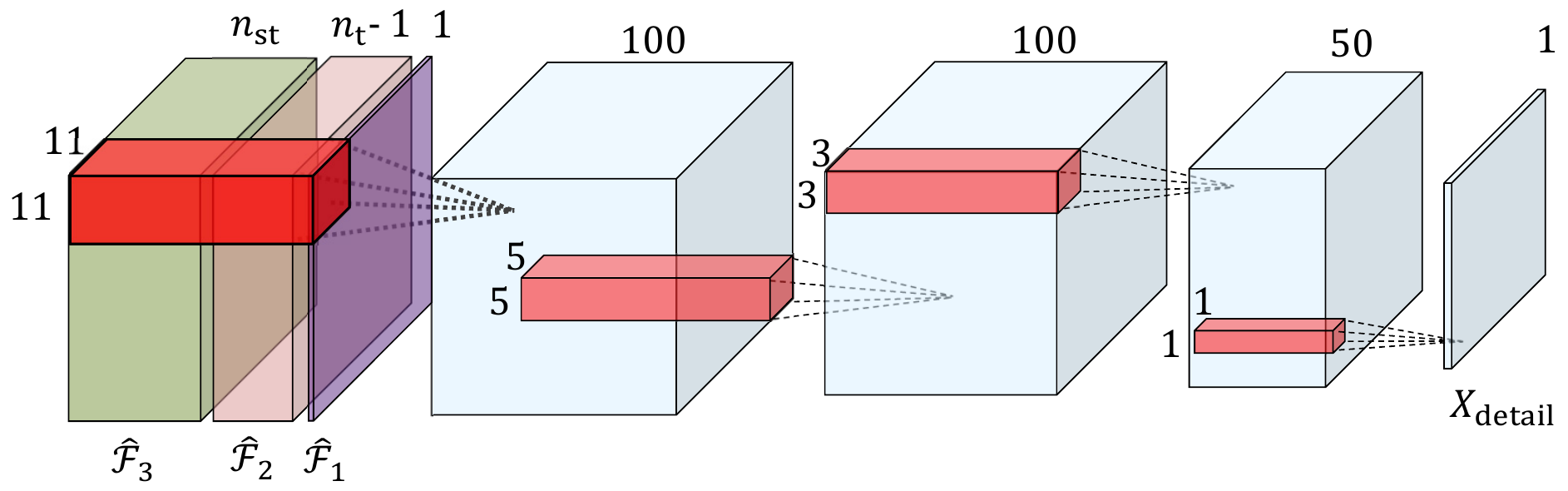}
	\end{center}
	\vspace{-0.5cm}
	\caption{CNN architecture for compensation of mis-alignment blur. Each convolutional layer is followed by a rectified linear unit (ReLU).}
	\vspace{-0.5cm}
	\label{fig_cnnArchitecture}
\end{figure}

\subsubsection{High Frequency Detail Feature} 
 
The matched slices in $\mathcal{T}_1$ are sorted according to their rain-free resemblance to $X_k$, which provide good reference to the content details with supposedly small mis-alignment. We directly use the tensor $\mathcal{T}_1$ as the last group of features $\mathcal{F}_3= \mathcal{T}_1\in\mathbb{R}^{n_x\times n_x\times n_{st}}$. This feature will compensate the detail loss introduced by the operations in Eq. (\ref{eqn_feature1}) for \textbf{false positive} rain pixels.

In order to facilitate the network training, we limit the mapping range between the input features and regression output by removing the low frequency component ($X_\text{avg}$) from these input features. Pixels in $X_k$ but outside of the SP $\mathcal{P}_k$ is masked out with $M_\text{SP}$:\vspace{-0.1cm}
\begin{align}
\hat{\mathcal{F}}_1&= (\mathcal{F}_1- X_\text{avg})\odot M_\text{SP},\\ \notag
\hat{\mathcal{F}}_2&= (\mathcal{F}_2- \mathcal{R}(X_\text{avg},n_t-1))\odot\mathcal{R}(M_\text{SP},n_t-1),\\ \notag
\hat{\mathcal{F}}_3&= (\mathcal{F}_3- \mathcal{R}(X_\text{avg},n_{st}))\odot\mathcal{R}(M_\text{SP},n_{st}).
\end{align}
The final input feature set is $\{\hat{\mathcal{F}}_1,~\hat{\mathcal{F}}_2,~\hat{\mathcal{F}}_3\}$. The feature preparation process is summarized in Fig. \ref{fig_T0T1forming}.

\subsubsection{CNN Structure and Training Details}\label{sec_cnn}

\begin{figure}
	\centering
	\includegraphics[width=0.95\linewidth]{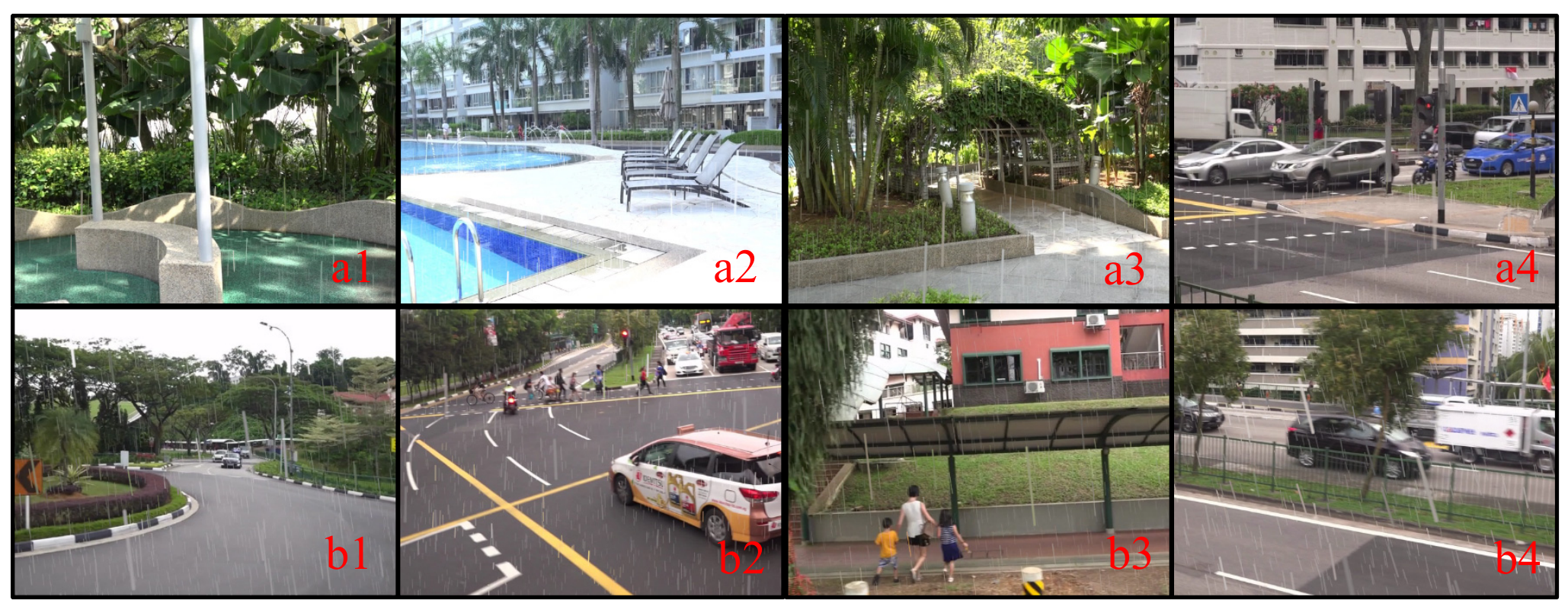}
	\caption{8 testing rainy scenes synthesized with \textit{Adobe After Effects} \cite{adobeAE}. First row (Group \textit{a}) are shot with a panning unstable camera. Second row (Group \textit{b}) are from a fast moving camera (speed range between 20 to 30 \textit{km/h})}
	\vspace{-0.5cm}
	\label{fig_testSet}
\end{figure}

\begin{figure*}[!t]
	\centerline{\subfloat{\includegraphics[width=1\linewidth]{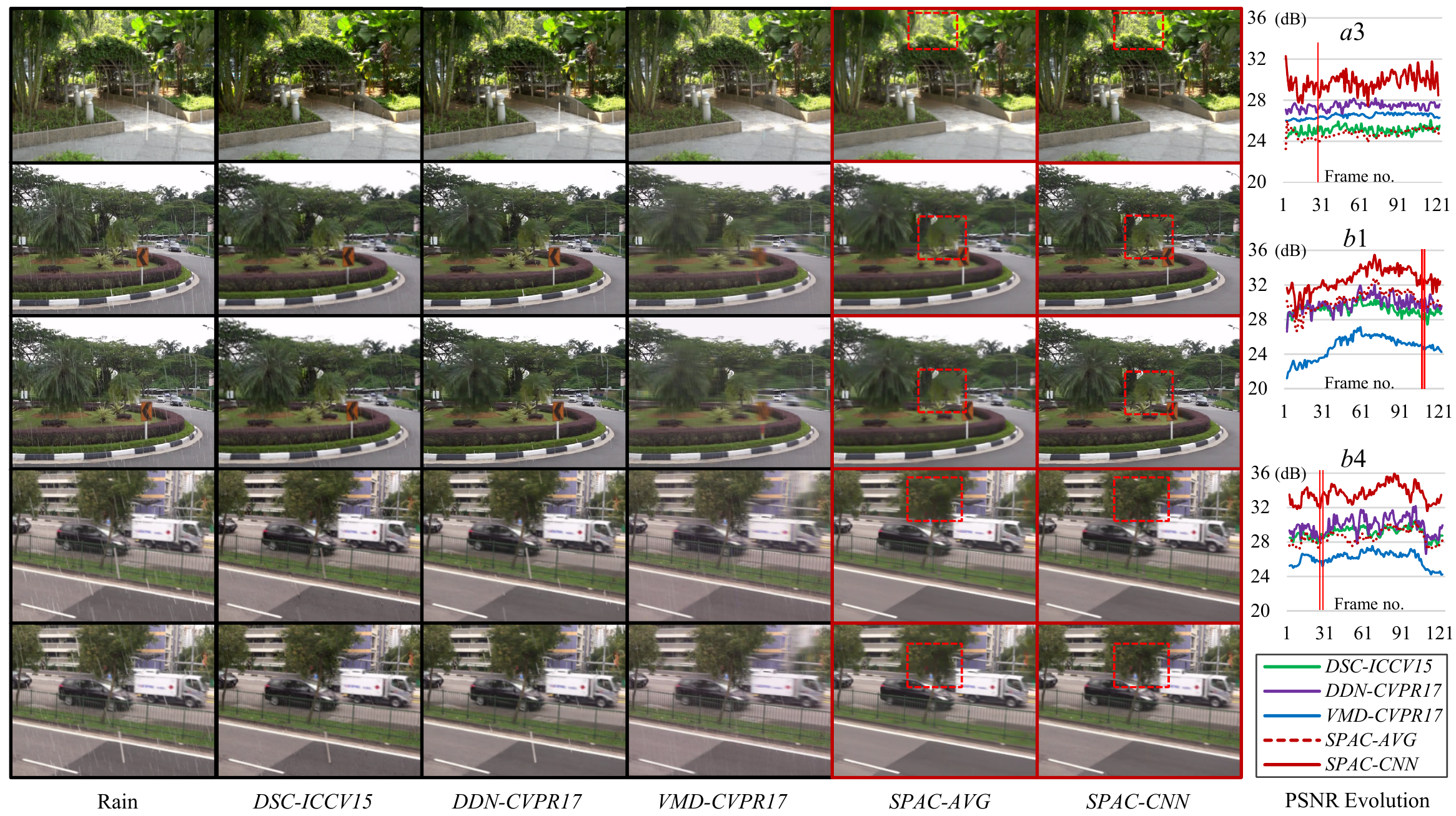}}}
	\caption{Visual comparison for different rain removal methods on the synthetic testing data \textit{a}3 (row 1), two consecutive frames for the data \textit{b}1 (row 2-3), and \textit{b}4 (row 4-5). The PSNR evolution curves for each frame are shown on the right.}
	\vspace{-0.2cm}
	\label{fig_visualComp_testSet}
\end{figure*}

\begin{table*}[ht]
	\small
	\centering\setlength\tabcolsep{1.8pt} 
	\begin{center}		
		\caption{Rain removal performance comparison between different methods in terms of scene reconstruction PSNR/SSIM, and F-measure for rain streak edge PR curves.} 
		\vspace{-0.3cm}
		\label{tbl_psnr}
		\begin{tabular}{|>{\centering\arraybackslash}m{1.1cm}|>{\centering\arraybackslash}m{0.8cm}|>{\centering\arraybackslash}m{0.8cm}|>{\centering\arraybackslash}m{0.8cm}|>{\centering\arraybackslash}m{0.6cm}|>{\centering\arraybackslash}m{0.8cm}|>{\centering\arraybackslash}m{0.8cm}|>{\centering\arraybackslash}m{0.6cm}|>{\centering\arraybackslash}m{0.8cm}|>{\centering\arraybackslash}m{0.8cm}|>{\centering\arraybackslash}m{0.6cm}|>{\centering\arraybackslash}m{0.8cm}|>{\centering\arraybackslash}m{0.8cm}|>{\centering\arraybackslash}m{0.6cm}|>{\centering\arraybackslash}m{0.8cm}|>{\centering\arraybackslash}m{0.8cm}|>{\centering\arraybackslash}m{0.6cm}|>{\centering\arraybackslash}m{0.8cm}|>{\centering\arraybackslash}m{0.8cm}|}
			\hline 
			\multirow{2}{*}{\parbox{1cm}{\textit{Camera\\Motion}}} &\multirow{2}{*}{\parbox{0.5cm}{\textit{Clip No.}}}　&\multicolumn{2}{c|}{Rain}&\multicolumn{3}{c|}{\textit{DSC-ICCV15} \cite{luo2015removing}} &\multicolumn{3}{c|}{\textit{DDN-CVPR17} \cite{fu2017removing}} &\multicolumn{3}{c|}{\textit{VMD-CVPR17} \cite{ren2017video}} &\multicolumn{3}{c|}{\textit{SPAC-Avg}} &\multicolumn{3}{c|}{\textit{SPAC-CNN}}\\
			\cline{3-19}
			&　&PSNR &SSIM &F &PSNR &SSIM  &F &PSNR &SSIM &F &PSNR &SSIM &F &PSNR &SSIM &F &PSNR &SSIM \\
			\hline\hline
			\multirow{5}{*}{\parbox{1cm}{\textit{panning\\unstable\\camera}}} 
			&\textit{a}1 &28.46 &0.94 &0.38 &25.61 &0.93 &0.47 &28.02 &0.95 &0.47 &26.96 &0.92 &0.39 &24.78 &0.87 &\textbf{0.51} &\textbf{29.78} &\textbf{0.97} \\ 
			&\textit{a}2 &28.09 &0.95 &0.33 &27.11 &0.95 &0.44 &27.38 &0.95 &\textbf{0.51} &24.80 &0.93 &0.40 &26.34 &0.89 &\textbf{0.51} &\textbf{30.09} &\textbf{0.96} \\
			&\textit{a}3 &27.84 &0.93 &0.43 &25.08 &0.92 &0.45 &27.41 &0.94 &0.42 &26.45 &0.90 &0.40 &24.72 &0.85 &\textbf{0.54} &\textbf{29.75} &\textbf{0.96} \\
			&\textit{a}4 &31.48 &0.95 &0.34 &28.82 &0.95 &0.53 &32.47 &0.97 &\textbf{0.55} &29.55 &0.94 &0.48 &29.90 &0.93 &0.54 &\textbf{34.82} &\textbf{0.98} \\\cline{2-19}
			&avg. \textit{a} &28.97 &0.94 &0.37 &26.66 &0.94 &0.47 &28.82 &0.95 &0.49 &26.94 &0.92 &0.42 &26.44 &0.89 &\textbf{0.53} &\textbf{31.11} &\textbf{0.97} \\ \hline\hline
			\multirow{5}{*}{\parbox{0.9cm}{\textit{camera\\speed\\20-30\\km/h}}} 
			&\textit{b}1 &28.72 &0.92 &0.42 &28.78 &0.92 &0.53 &29.48 &0.96 &0.35 &24.09 &0.84 &0.47 &26.35 &0.89 &\textbf{0.55} &\textbf{31.19} &\textbf{0.96} \\
			&\textit{b}2 &29.49 &0.90 &0.43 &29.58 &0.92 &0.50 &30.23 &0.95 &0.43 &25.81 &0.89 &0.50 &28.83 &0.93 &\textbf{0.57} &\textbf{34.05} &\textbf{0.98} \\
			&\textit{b}3 &31.04 &0.95 &0.33 &29.55 &0.95 &\textbf{0.53} &31.39 &0.97 &0.43 &26.12 &0.90 &0.48 &29.55 &0.94 &\textbf{0.53} &\textbf{33.73} &\textbf{0.98} \\
			&\textit{b}4 &27.99 &0.92 &0.50 &29.10 &0.93 &0.51 &29.83 &0.96 &0.48 &25.90 &0.88 &0.53 &28.85 &0.92 &\textbf{0.58} &\textbf{33.79} &\textbf{0.97} \\\cline{2-19}
			&avg. \textit{b} &29.31 &0.92 &0.42 &29.25 &0.93 &0.52 &30.23 &0.96 &0.42 &25.48 &0.88 &0.50 &28.40 &0.92 &\textbf{0.56} &\textbf{33.19} &\textbf{0.97} \\\hline 
		\end{tabular}
	\end{center}
	\vspace{-0.6cm}
\end{table*}

The CNN architecture is designed as shown in Fig. \ref{fig_cnnArchitecture}. The network consists of four convolutional layers with decreasing kernel sizes of 11, 5, 3, and 1. All layers are followed by a rectified linear unit (ReLU). Our experiments show this fully convolutional network is capable of extracting useful information from the input features and efficiently providing reliable predictions of the content detail $X_\text{detail}\in\mathbb{R}^{n_x\times n_x\times 1}$. The final rain removal output will be:\vspace{-0.2cm}
\begin{equation}
X_\text{derain}= X_\text{avg}+ X_\text{detail}.
\end{equation}

For the CNN training, we minimize the $\mathcal{L}_2$ distance between the derain output and the ground truth scene:
\begin{equation}\vspace{-0.2cm}
E=[\hat{X}- X_\text{avg}- X_\text{detail}]^2,
\end{equation}
here $\hat{X}$ denotes the ground truth clean image. We use stochastic gradient descent (SGD) to minimize the objective function. Mini-batch size is set as 50 for better trade-off between speed and convergence. The Xavier approach \cite{glorot2010understanding} is used for network initialization, and the ADAM solver \cite{kingma2014adam} is adpatoed for system training, with parameter settings $\beta_1=$ 0.9, $\beta_2=$ 0.999, and learning rate $\alpha=$ 0.0001.

To create the training rain dataset, we first took a set of 8 rain-free VGA resolution video clips of various city and natural scenes. The camera was of diverse motion for each clip, e.g.,  panning slowly with unstable movements, or mounted on a fast moving vehicle with speed up to 30 \textit{km/h}.
Next, rain was synthesized over these video clips with the commercial editing software \textit{Adobe After Effects} \cite{adobeAE}, which can create realistic synthetic rain effect for videos with adjustable parameters such as raindrop size, opacity, scene depth, wind direction, and camera shutter speed. This provides us a diverse rain visual appearances for the network training.

We synthesized 3 to 4 different rain appearances with different synthetic parameters over each video clip, which provides us 25 rainy scenes. For each scene, 21 frames were randomly extracted (together with their immediate buffer window for calculating features). Each scene was segmented into approximately 300 SPs, therefore finally we have around 157,500 patches in the training dataset.

\section{Performance Evaluation}\label{sec_experiments}

We set the sliding video buffer window size $n_t=5$. Each VGA resolution frame was segmented into around 300 SPs using the SLIC method \cite{achanta2012slic}. The bounding box size was $n_x=$ 80, and the spatial-temporal buffer $\mathcal{B}_k$ dimension was $n_\text{s}\times n_\text{s}\times n_\text{t}=$ 30$\times$30$\times$5. MatConvNet \cite{Vedaldi2015mat} was adopted for model training, which took approximately 54 hours to converge over the training dataset introduced in Sec. \ref{sec_cnn}. The training and all subsequent experiments were carried out on a desktop with Intel E5-2650 CPU, 56GB RAM, and NVIDIA GeForce GTX 1070 GPU.

\begin{figure*}[!t]
	\centerline{\subfloat{\includegraphics[width=0.9\linewidth]{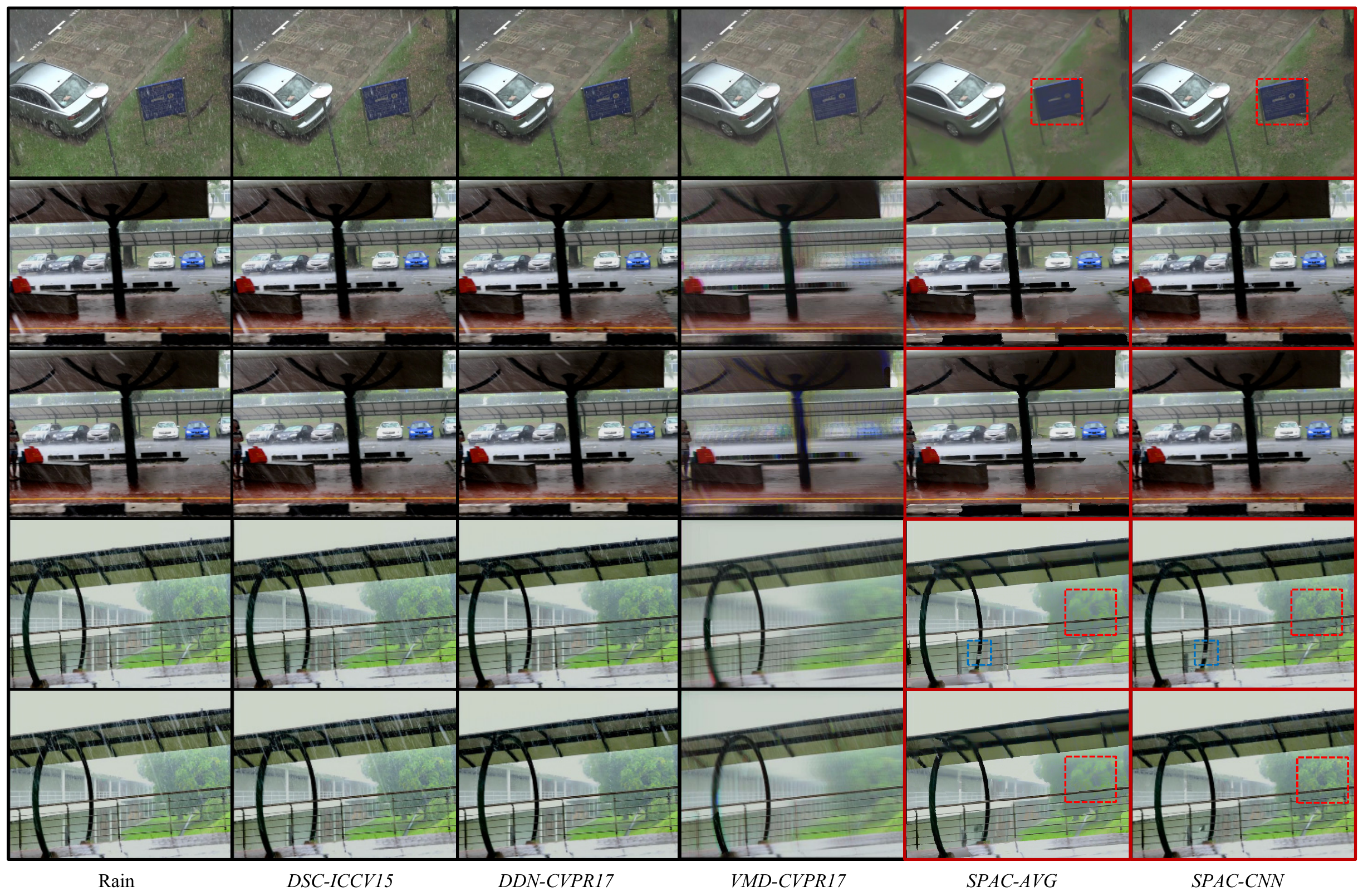}}}
	\vspace{-0.2cm}
	\caption{Visual comparison for different rain removal methods on real world rain data.}
	\vspace{-0.5cm}
	\label{fig_visualComp_realrain}
\end{figure*} \vspace{-0.2cm}

\begin{figure}
	\centering
	\includegraphics[width=1\linewidth]{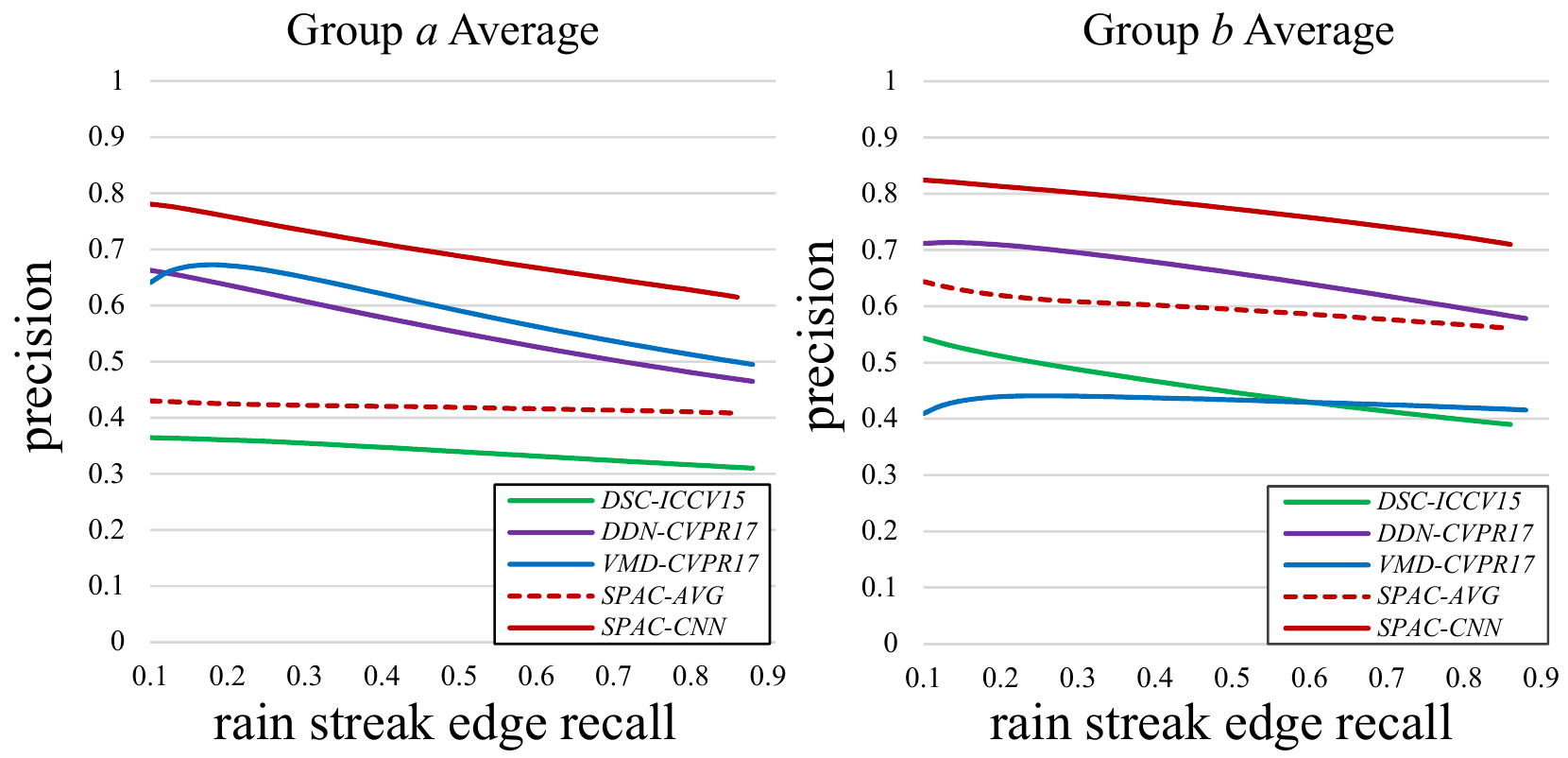}
	\vspace{-0.6cm}
	\caption{Rain edge pixel detection precision-recall curves for different rain removal methods.}
	\vspace{-0.6cm}
	\label{fig_prCurves}
\end{figure}
 
\subsection{Quantitative Evaluation}

To quantitatively evaluate our proposed algorithm, we took a set of 8 videos (different from the training set), and synthesized rain over these videos with varying parameters. Each video is around 200 to 300 frames. All subsequent results shown for each video are the average of all frames. 

To test the algorithm performance in handling cameras with different motion, we divided the 8 testing scenes into two groups: Group \textit{a} consists of scenes shot from a panning and unstable camera; Group \textit{b} from a fast moving camera (with speed range between 20 to 30 \textit{km/h}). Thumbnails and the labeling of each testing scene are shown in Fig. \ref{fig_testSet}.

Four state-of-the-art methods were chosen for comparison: two image-based derain methods, i.e., discriminative sparse coding (\textit{DSC}) \cite{luo2015removing}, and the deep detail network (\textit{DDN}) \cite{fu2017removing}; one video-based method via matrix decomposition (\textit{VMD}) \cite{ren2017video}. The intermediate derain output $X_\text{Avg}$ is also used as a baseline (abbr. as  \textit{SPAC-Avg}). \vspace{-0.2cm}

\subsubsection{Rain Streak Edge Precision Recall Rates}\label{sec_pr}

Rain fall introduces edges and textures over the background. 
To evaluate how much of the modifications from the derain algorithm contributes positively to \textit{\textbf{only removing the rain pixels}}, we calculated the rain streak edge precision-recall (PR) curves.
Absolute difference values were calculated between the derain output against the scene ground truth. Different threshold values were applied to retrieve a set of binary maps, which were next compared against the ground truth rain pixel map to calculate the precision recall rates. 

Average PR curves for the two groups of testing scenes by different algorithms are shown in Fig. \ref{fig_prCurves}. As can be seen, for both Group \textit{a} and \textit{b}, \textit{SPAC-CNN} shows consistent advantages over \textit{SPAC-Avg}, which proves that the CNN model can efficiently compensate scene content details and suppress influences from rain streak edges.

Video-based derain methods (i.e., \textit{VMD} and \textit{SPAC-CNN}) perform better than image-based methods (i.e., \textit{DSC} and \textit{DDN}) for scenes in Group \textit{a}. With slow camera motion, temporal correspondence can be accurately established, which brings great advantage to video-based methods. 
However, with fast camera motion, the performance of \textit{VMD} deteriorates seriously for Group \textit{b} data: rain removal is now at the cost of background distortion. Image-based methods show its relative efficiency in this scenario. However, \textit{SPAC-CNN} still holds advantage over image-based methods at all recall rates for Group \textit{b} data, which shows its robustness for fast moving camera. \vspace{-0.2cm}

\subsubsection{Scene Reconstruction PSNR/SSIM}

We calculated the reconstruction PSNR/SSIM between different state-of-the-art methods against the ground truth, and the results are shown in Table \ref{tbl_psnr}. The F-measure for rain streak edge PR curves are also listed for each data. 

As can be seen, \textit{SPAC-CNN} is consistently 5 \textit{dB} higher than \textit{SPAC-Avg} for both Groups \textit{a} and \textit{b}. SSIM is also at least 0.06 higher. This further validates the efficiency of the CNN detail compensation network.

Video based methods (\textit{VMD} and \textit{SPAC-CNN}) show great advantages over image-based methods for Group \textit{a} data (around 2dB and 5dB higher respectively than \textit{DSC}). For Group \textit{b}, image-based methods excel \textit{VMD}, however \textit{SPAC-CNN} still hold a 3dB advantage over \textit{DDN}, 4dB over \textit{DSC}. \vspace{-0.7cm}

\subsubsection{Feature Evaluation}

We evaluated the roles different input features play in the final derain PSNR over two testing data \textit{a}1 and \textit{b}4. Three baseline CNNs with different combinations of features as input were independently trained for this evaluation. As can be seen from the results in Table. \ref{tbl_evalFeatures}, combination of the three features $\hat{\mathcal{F}}_1+\hat{\mathcal{F}}_2+\hat{\mathcal{F}}_3$ provides the highest PSNR. $\mathcal{F}_1$ proves to be the most important feature. Visual inspection on the derain output show both $\hat{\mathcal{F}}_2$+$\hat{\mathcal{F}}_3$ and $\hat{\mathcal{F}}_1$+$\hat{\mathcal{F}}_3$ leaves significant amount of un-removed rain. Comparing the last two columns, it can be seen that  $\hat{\mathcal{F}}_3$ works more efficiently with \textit{a}1 than \textit{b}4, which makes sense since the high frequency features are better aligned for slow cameras, which led to more accurate detail compensation.

\begin{table}[t]
	\begin{center}
		\small\centering\setlength\tabcolsep{1pt}
		\caption{Derain PSNR (\textit{dB}) with different features absent.}
		\vspace{-0.2cm} 
		\label{tbl_evalFeatures}
		\begin{tabular}{|>{\centering\arraybackslash}m{0.8cm}|>{\centering\arraybackslash}m{1.3cm}|>{\centering\arraybackslash}m{1.3cm}|>{\centering\arraybackslash}m{1.3cm}|>{\centering\arraybackslash}m{1.8cm}|}
			\hline 
			\textit{\parbox{0.6cm}{Clip \\ No.}} &\parbox{1.2cm}{$\hat{\mathcal{F}}_2$+$\hat{\mathcal{F}}_3$ \\ (w/o $\hat{\mathcal{F}}_1$)} &\parbox{1.2cm}{$\hat{\mathcal{F}}_1$+$\hat{\mathcal{F}}_3$ \\ (w/o $\hat{\mathcal{F}}_2$)} &\parbox{1.2cm}{$\hat{\mathcal{F}}_1$+$\hat{\mathcal{F}}_2$ \\ (w/o $\hat{\mathcal{F}}_3$)} &\parbox{1.5cm}{$\hat{\mathcal{F}}_1$+$\hat{\mathcal{F}}_2$+$\hat{\mathcal{F}}_3$}\\ \hline \hline 
			\textit{a}1 &25.28 &28.87 &27.63 &\textbf{29.78}\\\hline
			\textit{b}4 &28.62 &31.97 &32.99 &\textbf{33.79}\\\hline			
		\end{tabular}
	\end{center}
	\vspace{-0.8cm}
\end{table}

\subsection{Visual Comparison}

We carried out visual comparison to examine the derain performance of different algorithms. Fig. \ref{fig_visualComp_testSet} shows the derain output for the testing data \textit{a}.3, \textit{b}.1, and \textit{b}.4. Two consecutive frames are shown for \textit{b}.1 and \textit{b}.4 to demonstrate the camera motion. As can be seen, image-based derain methods can only handle well light and transparent rain occlusions. For those opaque rain streaks that cover a large area, they fail unavoidably. Temporal information proves to be critical in truthfully restoring the occluded details.

It is observed that rain can be much better removed by video-based methods. However the \textit{VMD} method creates serious blur when the camera motion is fast. The derain effect for \textit{SPAC-CNN} is the most impressive for all methods. The red dotted rectangles showcase the restored high frequency details between \textit{SPAC-CNN} and \textit{SPAC-Avg}.

Although the network has been trained over synthetic rain data, experiments show that it generalizes well to real world rain. Fig. \ref{fig_visualComp_realrain} shows the derain results. As can be seen, the advantage of \textit{SPAC-CNN} is very obvious under heavy rain, and robust to fast camera motion.\vspace{-0.2cm}

\subsection{Execution Efficiency}

\begin{table}[t]
	\begin{center}
		\small\centering\setlength\tabcolsep{1pt}
		\caption{Execution time (in \textit{sec}) comparison for different methods on deraining a single VGA resolution frame.}
		\vspace{-0.2cm} 
		\label{tbl_runtime}
		\begin{tabular}{|>{\centering\arraybackslash}m{1.5cm}|>{\centering\arraybackslash}m{1.5cm}|>{\centering\arraybackslash}m{1.5cm}|>{\centering\arraybackslash}m{1.5cm}|>{\centering\arraybackslash}m{1.5cm}|}
			\hline 
			\multirow{2}{*}{\textit{DSC} \cite{luo2015removing}} &\multirow{2}{*}{\textit{DDN} \cite{fu2017removing}} &\multirow{2}{*}{\textit{VMD} \cite{ren2017video}} &\multicolumn{2}{c|}{\textit{SPAC-CNN}}\\\cline{4-5}
			& & &\textit{SPAC-Avg} &CNN\\\hline\hline
			Matlab &Matlab &Matlab &C++ &Matlab \\ \hline
			236.3 &0.9 &119.0 &0.2 &3.1\\\hline
		\end{tabular}
	\end{center}
	\vspace{-0.8cm}
\end{table}

We compared the average runtime between different methods for deraining a VGA resolution frame. Results are shown in Table \ref{tbl_runtime}. As can be seen \textit{SPAC-Avg} is much faster than all other methods. \textit{SPAC-CNN} is much faster than video-based method, and it's comparable to that of \textit{DDN}. \vspace{-0.5cm}

\section{Discussion}\label{sec_limitations}

For \textit{SPAC-CNN}, the choice of SP as the basic operation unit is key to its performance. When other decomposition units are used instead (e.g., rectangular), matching accuracy deteriorates, and very obvious averaging blur will be introduced especially at object boundaries.

Although the SP template matching can only handle translational motion, alignment errors caused by other types of motion such as rotation, scaling, and non-ridge transforms can be mitigated with global frame alignment before they are buffered (as shown in Fig. \ref{fig_system}) \cite{tan2014dynamic}. Furthermore, these errors can be efficiently compensated by the CNN.

When camera moves even faster, SP search range $n_s$ needs to be enlarged accordingly, which increases computation loads.
We have tested scenarios with camera speed going up to 50 \textit{km/h}, the PSNR becomes lower due to larger mis-alignment blur, alignment error is also possible as showcased in blue rectangles in Fig. \ref{fig_visualComp_realrain}. We believe a re-trained CNN with training data from such fast moving camera will help improve the performance. \vspace{-0.2cm} 

\section{Conclusion}\label{sec_conclusion}

We have proposed a video-based rain removal algorithm that can handle torrential rain fall with opaque streak occlusions from a fast moving camera. 
SP have been utilized as the basic processing unit for content alignment and occlusion removal. A CNN has been designed and trained to efficiently compensate the mis-alignment blur introduced by deraining operations. The whole system shows its efficiency and robustness over a series of experiments which outperforms state-of-the-art methods significantly.\vspace{-0.2cm}


\section*{Acknowledgment}
\addcontentsline{toc}{section}{Acknowledgment}
The research was partially supported by the ST Engineering-NTU Corporate Lab through the NRF corporate lab@university scheme.

{\small
\bibliographystyle{ieee}
\bibliography{mybib}
}

\end{document}